# Emergent symbolic language based deep medical image classification

Aritra Chowdhury, Alberto Santamaria-Pang, James R. Kubricht, Peter Tu

Artificial Intelligence, GE Research, 1 Research Circle, Niskayuna NY 12309, USA

**Abstract.** Modern deep learning systems for medical image classification have demonstrated exceptional capabilities for distinguishing between image based medical categories. However, they are severely hindered by their inability to explain the reasoning behind their decision making. This is partly due to the uninterpretable continuous latent representations of neural networks. Emergent languages (EL) have recently been shown to enhance the capabilities of neural networks by equipping them with symbolic representations in the framework of referential games. Symbolic representations are one of the cornerstones of highly explainable good old fashioned AI (GOFAI) systems. In this work, we demonstrate for the first time, the emergence of deep symbolic representations of emergent language in the framework of image classification. We show that EL based classification models can perform as well as, if not better than state of the art deep learning models. In addition, they provide a symbolic representation that opens up an entire field of possibilities of interpretable GOFAI methods involving symbol manipulation. We demonstrate the EL classification framework on immune cell marker based cell classification and chest X-ray classification using the *CheXpert* dataset. Code is available online at `https://github.com/AriChow/EL`.

## 1 Introduction

The recent deep learning revolution began with the seminal AlexNet [1] paper that used convolution neural networks (CNN) to perform image recognition. This led to the development of networks like ResNet [2], InceptionNet [3] and MobileNet [4] among others. The medical imaging community soon followed suit in adopting said methods and deep learning approaches to tackle the medical image classification task. Surprisingly, results showed [5] that methods like transfer learning and fine tuning had capabilities of performing exceedingly well on medical image analysis tasks. This relatively unintuitive method led to the explosion of transfer learning based approaches using CNNs trained on large image corpuses and then being fine-tuned on the target medical dataset. The results exceeded expectations. Traditional methods involving painstakingly curated features developed using human intuition had now taken a backseat with the arrival of monster sized deep networks that doubled and tripled classification accuracies overnight. It was at that opporune moment that such inscrutable methods hit a roadblock. Medical practitioners and policy makers refused to adopt black box deep learning methods unless they are able to explain the logic behind their decision making. Interpretability was the need of the hour.

Symbolic methods have been around in the field of artificial intelligence for over half a century. They unfortunately suffered from being too inflexible and were hard to generalize over large populations of data. These methods were eventually sacrificed in favor of connectionist approaches based on continuous latent representations.

Our novel approach of combining symbols with connectionist deep networks creates a new direction of bringing back symbolic methods in from the cold. We seem to have come full circle.

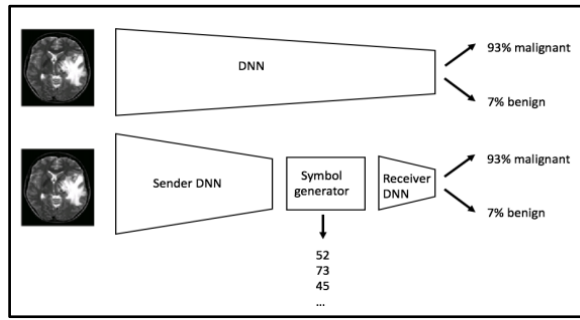

**Fig. 1.** Emergent symbolic classification (bottom) as compared to the traditional CNN framework (top). The emergent symbolic framework consists of a sender network, a symbol generator and a receiver network.

We introduce a new paradigm of image classification, in particular medical image classification as shown in Fig. 1. Our approach combines the representative power of deep learning and the interpretable capabilities of symbols. Specifically, we use the framework of emergent languages to generate symbolic representations in the context of medical image recognition.

The idea of emergent languages is based on research involving multi-agent communication games where a sender and receiver coordinate to solve a task or game. The original work involved solving a variant of referential games known as *Lewis signaling game* [6]. Our contribution lies in introducing this method in the framework of medical image classification.

We demonstrate our methodology on two tasks. The first task is a cell classification game which involves classification of features using a fully connected deep neural network (DNN) into 4 classes of protein markers – *CD3*, *CD20*, *CD68* and *Claudin1*. The second task involves classification of Chest X-rays of patients into two categories – indication of *pleural effusion* and *no* indication of *pleural effusion*. This dataset is a subset of the CheXpert [7] dataset. A CNN is used to perform classification on this task. Our experiments show that our EL based medical image classification methodology performs as well as, if not better in terms of classification accuracy on the test dataset. More importantly, the availability of symbolic representations provides an avenue for interpreting these complex networks and explain the logic behind their decision making.

## 2      Prior work

Medical image classification has been an active field of research in the medical image community among multiple modalities including CT, MRI, Ultrasound, PT and endoscopy among others. Traditionally, methods like support vector machines and random forests were used to perform classification over hand-crafted features [8]. The advent of deep learning has led to the development of new methods based on convolutional neural networks for performing image classification [9]. We introduce a modification of the CNN that allows us to extract symbolic representations.

Numerous approaches for model interpretability have been proposed recently. They include visualization of CNN representations [10], interpreting gradients [11], and perturbation based methods [12]. Our approach to interpretability is through the introduction of symbolic representations that maybe used for downstream symbolic manipulations and induction of logical rules.

In this work we introduce the framework of emergent languages in image classification. This framework is inspired by Lazaridou et. al [13], where they introduce the idea of using referential games for multi-agent cooperation and show emergence of artificial language. They also discuss ideas to ground the symbols in natural languages. In this work, we make an attempt to formulate and extend the ideas of emergent language communication to come up with a method for generating symbols to perform classification.

## 3      Data

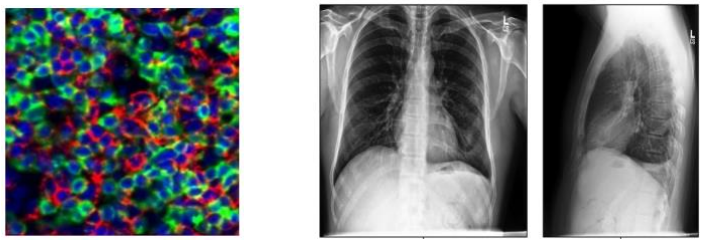

**Fig. 2.** (Left) Color images for immune cell markers for *CD20* (red), *CD3* (green), and *Nuclei DAPI* (blue). (Right) Frontal and lateral views of chest X-rays from *CheXpert* dataset.

### 3.1      Immunofluorescence cell markers

We use a colon cohort in this analysis whose collection methods and details are provided in [14]. In this work, we use 4 cell markers – *CD3, CD20, CD68* and *Claudin1*. Seven statistical intensity and shape based features are extracted from each cell marker encoding the information relevant to each marker. This makes it a total of 28 features

– 1 set of 7 features for each of the 4 markers. We divide this dataset into 534 samples of training data, 133 samples of validation data, and 171 samples of test data.

### 3.2 CheXpert pleural effusion

This dataset is a subset of the *CheXpert* dataset [7]. The original dataset consists of images with 14 different indications. It consists of 224316 radiographs of 65240 patients with frontal and lateral views of the lungs as shown in Fig. 2. The training dataset consists of ground truth of the 14 observations corresponding to different diseases. For this work, we used a subset of the data consisting of existence and absence of indications of pleural effusion. We divide this dataset into 97265 samples of training data, 24318 samples of validation data, 234 and samples of test data.

## 4 Methods

### 4.1 Emergent symbolic classification

Our primary contribution involves the introduction of emergent languages to the classification framework as shown in Fig. 1. The basic setup involves a sender architecture, a symbol generator and receiver architecture. The sender can be any network that extracts feature representations from input data. The sender sends the feature representations to the symbol generator where symbols are generated. These symbols are then fed to a receiver network that performs the classification. The only information that flows from the sender to the receiver are discrete representations instead of continuous features. The whole network involving the sender, generator and receiver networks are trained end-to-end. In our work, we show that just by using one symbol, the sender and receiver architectures are able to communicate to solve the task of medical image classification. The symbol generator is approximated as a Gumbel softmax estimator [15]. Sampling is done on the sender input using this method in order to ensure that the network remains fully differentiable. One hot encoded symbols $w \in V$, where $V$ is the vocabulary of the symbols, are sampled from a categorical distribution using a continuous relaxation $\widetilde{w}$ is obtained from the Gumbel-softmax sampling given by the following equation.

$$\widetilde{w}_k = \frac{\exp((\log p_k + g_k)/\tau)}{\sum_{i=1}^{K} \exp((\log p_i + g_i)/\tau)} \quad (1)$$

where, $K$ is the number of samples, $p_i$ are the probabilities of the categorical distribution, $\tau$ is the temperature that controls the accuracy of the approximation. $g_i$ is formulated as follows.

$$g_i = -\log(-\log(u_i)) \quad (2)$$

where, $\{u_i\}_{i=1}^{K}$ is sampled from a uniformly distributed variable $u \sim U(0,1)$. This relaxation makes the symbol generator completely differentiable.

### 4.2 Neuron conductance

This method [16] is used to study the attribution of the input features with respect to the symbols generated by the symbol generator. This basically represents the importance of a hidden unit to the prediction over a set of inputs. It is given by the following equation.

$$Cond^y(x) ::= \sum_i (x_i - x'_i) \cdot \int_{\alpha=0}^{1} \frac{\partial F(x' + \alpha(x-x'))}{\partial y} \cdot \frac{\partial y}{\partial x_i} \partial \alpha \qquad (3)$$

where, $x_i$ is the input and $x'_i$ is the baseline, $y$ is the hidden neuron, $F: R^n \to [0, 1]$ is the function that represents the deep network. The conductance of a particular neuron builds on Integrated Gradients (IG) [17] by looking at the flow of IG attribution from each input through the particular neuron.

## 5 Experiments and results

### 5.1 Experimental setup

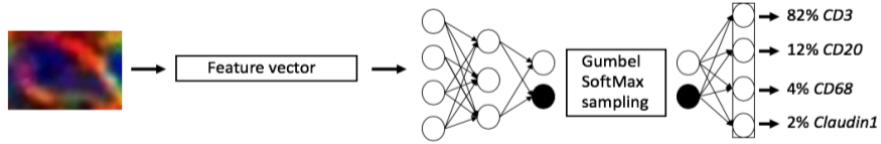

**Fig. 3.** Experimental setup for immune cell marker classification. This consists of fully connected neural networks for sender and receiver architectures.

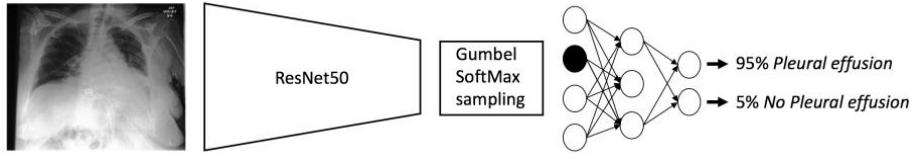

**Fig. 4.** Experimental setup for *CheXpert* pleural effusion classification. The sender architecture is a ResNet50 CNN and receiver architecture is a fully connected network that takes as input the symbolic representations from the Gumbel softmax sampler.

**Immune cell marker classification.** The experimental setup experiment (referred as '*Immune cell* symbol' in experiments) of the immune cell marker classification is described graphically in Fig. 3. The dataset used is described in Section 3.1. Quantitative features are extracted from the images of the immune cell markers. These features have a dimensionality of 28. They are fed to a fully connected sender network consisting of 2 hidden layers. The output of the last hidden layer is sent to the Gumbel-softmax sampler that generates a symbol from a vocabulary of size 100. This symbolic representation is sent to the receiver which is a 2-layer DNN that performs the classification into one of the 4 protein markers – *CD3, CD20, CD68* and *Claudin1*.

**CheXpert pleural effusion.** This experiment (referred as '*CheXpert* symbol' in experiments) is setup along the same lines as above except that the sender architecture is a CNN, in particular the ResNet50 architecture. The penultimate 2048 dimensional feature representation is extracted from the CNN and fed to the sampler. The sampler then generates a symbol from a vocabulary of size 100, which is then fed to the receiver that has a similar architecture as the immune cell marker experiment. The output softmax is then used to classify between existence and absence of *pleural effusion* as shown in Fig. 4.

Both the experiments are compared to the baseline traditional classification methodology denoted in Fig. 1. The control of the immune cell marker experiment is basically the same setup as Fig. 3, without the Gumbel-softmax sampler. The sender input is directly fed to the receiver inputs. This is referred to as '*Immune cell* baseline' in Table 1. The control for the *CheXpert* experiment is similar. The continuous 2048 dimensional continuous representation from the sender ResNet is forwarded to the receiver DNN. This is referred to as '*CheXpert* baseline' in Table 1.

**Implementation details.** The parameters of the networks were set to be consistent over the different experiments including the baseline. The optimizer used was the Adam optimizer [18] with a learning rate of *1e-3*. Cross entropy was used as the loss function. The batch size was set as 32 and the model used for testing was generated by using early stopping on the validation loss. The vocabulary size of both experiments were set to 100. Pytorch [19] was the framework used for performing the experiments. The implementations of the gumbel softmax sampling is inspired from [20]. The implementation is available online at `https://github.com/AriChow/EL`.

## 5.2 Results

**Table 1.** Comparison of performance statistics between the baseline control experiments and the *CheXpert* and *Immune cell* symbolic classifiers. The symbols generated are also reported.

| Experiment | Accuracy (%) | F1-score | Symbols |
|---|---|---|---|
| *CheXpert* baseline | 73.07 | 0.677 | None |
| *CheXpert* symbol | 77.35 | 0.697 | 20, 66, 73, 85, 88 |
| *Immune cell* baseline | 98.20 | 0.982 | None |
| *Immune cell* symbol | 97.07 | 0.9713 | 4, 20, 47, 58 |

**Comparison of classification performance with baselines.** Table 1 shows a comparison of the experiments with respect to their respective controls. These results are on a held-out test dataset that consists of 171 samples for the immune cell marker experiment and 234 samples for the *CheXpert* experiment. We observe that the performance of the emergent symbolic classifier for the immune cell marker classification is comparable with the control experiment with a slight drop in performance. In the case of the *CheXpert* experiments, the emergent symbolic classifier actually performs better than the baseline experiment both in terms of accuracy and F1-score. One of the reasons for such a behavior is that the sampling introduces a kind of bottleneck in the network and this acts as a regularizer, thus enhancing the generalization capabilities of the symbolic CNN. This shows that a symbol based deep learning method can be used instead of a black box CNN. We also observe that the number of unique symbols is very small compared to the vocabulary. This behavior has been studied in terms of information theory recently [21].

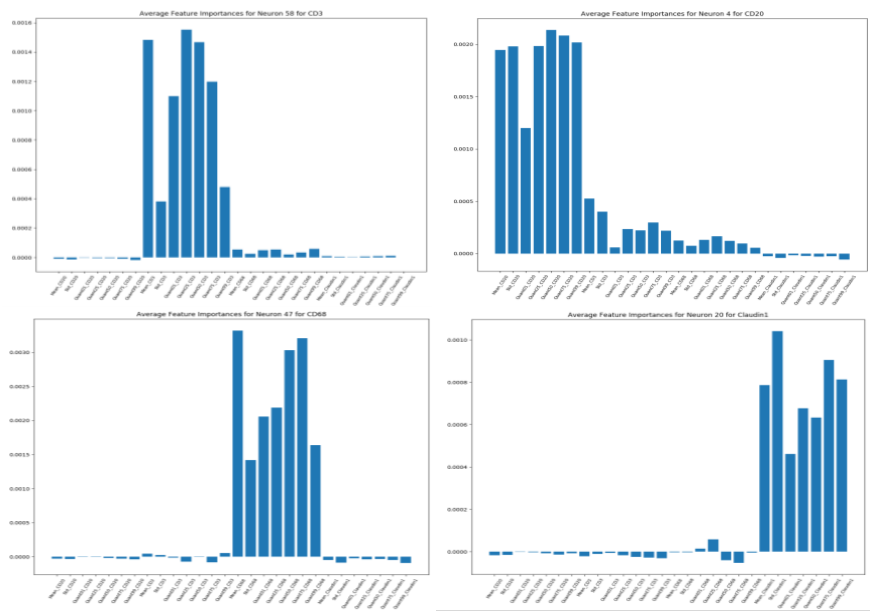

**Fig. 5.** Average neuron conductance attributions of each symbol with respect to the 28 dimensional input features. The x-axis represents the features. The y-axis represents the attributions using Eq. 3.

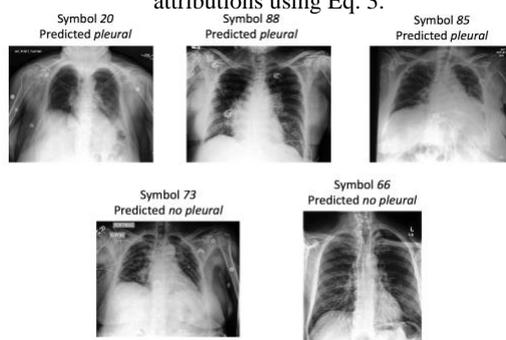

**Fig. 6.** Chest X-ray samples from test dataset corresponding to the predicted symbols.

**Interpretability of protein marker features using neuron conductance.** We extract the neuron conductance values of each of the symbols with respect to the 28 dimensional input features. Fig. 5 shows the distribution of the attributions or feature importance with respect to the symbols corresponding to each marker in Table 1. The values are computed over the 171 test samples and averaged. The input features consist of 7 quantitave features corresponding to the 4 markers, in the order of *CD20*, *CD3*, *CD68* and *Claudin1*.The activations of the neuron conductance values verify that the symbols actually pick up relevant information representative of the features corresponding to the marker in the input feature vectors.

**Interpretability of Chest X-rays with respect to generated symbols.** We show the images of the *CheXpert* test datasets with respect to each of the symbols in Fig. 6. We observe that multiple symbols emerge corresponding to each of the 2 classes in the *CheXpert* dataset. Symbols 20, 88 and 85 correspond to predictions of *pleural effusion.* Symbols 66 and 73 correspond to predictions of *no* indications of *pleural effusion*.
We can also observe that each of the symbols correspond to different levels of phenotypical expressions in the input. For example, even though symbol 73 correctly indicates the non-existence of pleural effusion, it may also indicate existence of occlusion in the image. Symbol 66 could represent an X-ray with no occlusion. Symbols 20 and 85 can signify different extents of effusion in the lung cavity. Symbol 88 is interesting because the prediction is incorrectly classified as pleural effusion even though the ground truth has no indication. This symbol could represent uncertainty in the model or a possible error in annotation. The existence of symbol 88 can therefore be used to correct incorrect annotations in an active learning setting or can be sent to a medical expert to perform further investigation.

## 6   Conclusion

We introduce a new paradigm of classification of medical images using convolutional neural networks as shown in Fig. 1. This involves a modification of the CNN which transforms the network into 3 parts – the sender network, the symbol generator and the receiver network. The sender network generates a feature representation that is fed to the symbol generator. The generator consists of a sampling layer that uses the Gumbel-softmax estimator to produce a continuous relaxation of a one-hot encoding of the input features. The receiver network receives the symbolic representation and outputs the softmax layers like that of a traditional CNN based classifier. This encapsulates the emergent symbolic classifier. We show that the emergent symbolic classifier is able to perform as well as the baseline DNN for immune cell marker-based classification. We demonstrate using neuron conductance, how the generated symbols can be traced back to attributions from the input features. We perform another experiment using a subset of the *CheXpert* dataset consisting of radiographs with pleural effusion. The symbolic classifier performs better than the CNN baseline in terms of accuracy. We observe that the generated symbols indicate finer details of the input images.
The single symbol framework can be extended to use long short-term memory networks (LSTMs) [22] to form sentences of the emergent language. This could be used to interface with natural language and text like radiology reports to add another layer of interpretability to the black box nature of DNNs. The symbols generated by the Gumbel-softmax estimator can become the cornerstone of bridging symbolic methods with continuous representations of connectionist methods like neural networks.

**References**

1. Krizhevsky, A., Sutskever, I., Hinton, G. E.: Imagenet classification with deep convolutional neural networks. In: Advances in Neural Information Processing Systems, pp. 1097-1105 (2012)
2. He, K., Zhang, X., Ren, S., Sun, J.: Identity mappings in deep residual networks. In*: European conference on computer vision*, pp. 630-645. Springer, Cham (2016, October).
3. Szegedy, C., Vanhoucke, V., Ioffe, S., Shlens, J., Wojna, Z: Rethinking the inception architecture for computer vision. In: Proceedings of the IEEE conference on computer vision and pattern recognition, pp. 2818-2826 (2016).
4. Howard, A. G., Zhu, M., Chen, B., Kalenichenko, D., Wang, W., Weyand, T., Adam, H.: Mobilenets: Efficient convolutional neural networks for mobile vision applications. arXiv preprint arXiv:1704.04861 (2017).
5. Shin, H. C., Roth, H. R., Gao, M., Lu, L., Xu, Z., Nogues, I., Summers, R. M.: Deep convolutional neural networks for computer-aided detection: CNN architectures, dataset characteristics and transfer learning. In: IEEE transactions on medical imaging 35(5), 1285-1298 (2016).
6. Lewis D.: Convention. Harvard University Press, Cambridge, MA (1969).
7. Irvin, J., Rajpurkar, P., Ko, M., Yu, Y., Ciurea-Ilcus, S., Chute, C., Seekins, J.: Chexpert: A large chest radiograph dataset with uncertainty labels and expert comparison. In Proceedings of the AAAI Conference on Artificial Intelligence, Vol. 33, pp. 590-597, (2019, July).
8. Miranda, E., Aryuni, M., Irwansyah, E.: A survey of medical image classification techniques. In: IEEE International Conference on Information Management and Technology, pp. 56-61, (2016, November).
9. Litjens, G., Kooi, T., Bejnordi, B. E., Setio, A. A. A., Ciompi, F., Ghafoorian, M., Sánchez, C. I.: A survey on deep learning in medical image analysis. In: Medical image analysis, 42, 60-88 (2017).
10. Zhang, Q. S., Zhu, S. C.: Visual interpretability for deep learning: a survey. Frontiers of Information Technology & Electronic Engineering, 19(1), 27-39 (2018).
11. Selvaraju, R. R., Cogswell, M., Das, A., Vedantam, R., Parikh, D., Batra, D.: Grad-cam: Visual explanations from deep networks via gradient-based localization. In: Proceedings of the IEEE international conference on computer vision, pp. 618-626 (2017).
12. Zeiler, M. D., Fergus, R.: Visualizing and understanding convolutional networks. In: European conference on computer vision, pp. 818-833, Springer, Cham (2014, September)..
13. Lazaridou, A., Peysakhovich, A., Baroni, M.: Multi-agent cooperation and the emergence of (natural) language. In: arXiv preprint arXiv:1612.07182 (2016).
14. Santamaria-Pang, A., Huangy, Y., Rittscher, J.: Cell segmentation and classification via unsupervised shape ranking. In: 10th IEEE International Symposium on Biomedical Imaging, pp. 406-409 (2013, April).
15. Jang, E., Gu, S., Poole, B.: Categorical reparameterization with gumbel-softmax. In: arXiv preprint arXiv:1611.01144 (2016).
16. Dhamdhere, K., Sundararajan, M., Yan, Q.: How important is a neuron?. In: arXiv preprint arXiv:1805.12233 (2018).
17. Sundararajan, M., Taly, A., Yan, Q.: Axiomatic attribution for deep networks. In: Proceedings of the 34th International Conference on Machine Learning Volume 70, pp. 3319-3328 (2017, August).
18. Kingma, D. P., Ba, J: Adam: A method for stochastic optimization. In: arXiv preprint arXiv:1412.6980 (2014).

19. H Paszke, A., Gross, S., Chintala, S., Chanan, G., Yang, E., DeVito, Z., Lerer, A.: Automatic differentiation in pytorch (2017).
20. Kharitonov, E., Chaabouni, R., Bouchacourt, D., Baroni, M.: EGG: a toolkit for research on Emergence of lanGuage in Games. In: arXiv preprint arXiv:1907.00852 (2019).
21. Chaabouni, R., Kharitonov, E., Dupoux, E., Baroni, M: Anti-efficient encoding in emergent communication. In: Advances in Neural Information Processing Systems, pp. 6290-6300. (2019).
22. Havrylov, S., & Titov, I.: Emergence of language with multi-agent games: Learning to communicate with sequences of symbols. In: Advances in neural information processing systems, pp. 2149-2159 (2017).